# Aspect-Based Sentiment Analysis using Local Context Focus Mechanism with DeBERTa

Tianyu Zhao, Junping Du, Zhe Xue, Ang Li, Zeli Guan

Beijing Key Lab of Intelligent Telecomm. Software and Multimedia,
Beijing Univ. of Posts and Telecomm., Beijing 100876, China

**Abstract.** Text sentiment analysis, also known as opinion mining, is research on the calculation of people's views, evaluations, attitude and emotions expressed by entities. Text sentiment analysis can be divided into text-level sentiment analysis, sentence-level sentiment analysis and aspect-level sentiment analysis. Aspect-Based Sentiment Analysis (ABSA) is a fine-grained task in the field of sentiment analysis, which aims to predict the polarity of aspects. The research of pre-training neural model has significantly improved the performance of many natural language processing tasks. In recent years, pre training model (PTM) has been applied in ABSA. Therefore, there has been a question, which is whether PTMs contain sufficient syntactic information for ABSA. In this paper, we explored the recent DeBERTa model (Decoding-enhanced BERT with disentangled attention) to solve Aspect-Based Sentiment Analysis problem. DeBERTa is a kind of neural language model based on transformer, which uses self-supervised learning to pre-train on a large number of original text corpora. Based on the Local Context Focus (LCF) mechanism, by integrating DeBERTa model, we purpose a multi-task learning model for aspect-based sentiment analysis. The experiments result on the most commonly used the laptop and restaurant datasets of SemEval-2014 and the ACL twitter dataset show that LCF mechanism with DeBERTa has significant improvement.

## 1    Introduction

Sentiment Analysis (SA) is an active field in Natural Language Processing, which involves the viewpoint of text [1][2]. With the development of science and Technology Society, more and more people like to express opinions and comments on products after consumption, which has formed a large number of comment texts [3][4]. There are lots of entity attributes in the comment texts. Aspect-Based Sentiment Analysis (ABSA) [5][6] aims to do the fine-grained sentiment analysis towards aspects, which is the sub-task of Sentiment Analysis. Traditional methods explore the general emotions of texts, but ABSA is a more challenging task because opinions can contain several aspects. Take the sentence "Its <u>size</u> is ideal and the <u>weight</u> is acceptable" for example, the customer had a positive sentiment for both size and weight. The task is to predict the emotions towards the underlined aspects. Generally, the main research line of ABSA involves the identification of various aspect-level sentiment elements, namely, aspect terms, aspect categories, opinion terms and sentiment polarities [7]. ABSA tasks include four tasks, which are called aspect term extraction (ATE), aspect category detection (ACD), opinion term extraction (OTE), and aspect sentimental classification (ASC).ASC task is designed to predict the sentiment polarity for a particular aspect within a sentence. In this paper, we only focus on ASC tasks.

Early ASC tasks were usually based on manually designed functions, such as term frequency [8]. In recent years, ASC tasks based on deep learning has attracted extensive attention. BERT-based deep neural language models are widely used for ABSA [9][10]. Bert explores the recent deep bidirectional encoder representation from transformers, which processes text considering the context on the left and right sides of the word in all layers [11]. Bert can generate more text semantic representations, in which each word is mapped to an embedded vector, which depends on the context in the sentence. Pre-training methods (PTMs) such as BERT have brought significant performance improvements of the ASC task. Next, by training the model longer, with bigger batches, over more data, RoBERTa (A Robustly Optimized BERT Pre-training Approach) can exceed the performance of the traditional BERT method [12]. In 2020, He et al. [13] introduced a new architecture for BERT-based language model training called Decoding-enhanced BERT with Disentangled Attention (DeBERTa). Two new technologies are used to improve the previous state-of-the-art PLMs: a disentangled attention mechanism, and an enhanced mask decoder.

In this paper, we present a method for Aspect-Based Sentiment Analysis using Local Context Focus (LCF) Mechanism with DeBERTa. LCF mechanism combines local context features and global context features to predict sentiment polarity of targeted aspect [14]. In this way, the model can discover the unknown aspects and pay more attention to local context words of specific aspect. It is very important for sentient analysis based on domain specific aspects. Because the words representing aspects and sentiments are position dependent in the viewpoint text, they are usually close to each other. LCF can compute local context features. By further studying the application of PTMs in ABSA tasks, we made fine-tuning and mechanism enhancement on it.

Therefore, the main contributions of this paper are as follows:

1. This paper adds the latest PTM to the LCF design model for the first time. DeBERTa improves previous state-of-the-art PLMs using two novel techniques: a disentangled attention mechanism, and an enhanced mask decoder. This significantly improves the performance of LCF design.

2. On the basis of preprocessing, a new mechanism is added to enhance the relationship between local context features and global context features. Contextual features make the model better predict the polarity in aspects of the target.

3. Through the improvement and adjustment of the model, the performance of the ASC task is significantly improved. We conducted experiments with three datasets, which is the laptop and restaurant datasets of SemEval-2014 and the ACL twitter dataset. The experimental results show that the model is enhanced to some extent on three datasets, especially the Restaurant dataset.

## 2　　Related Works

ABSA tasks include four tasks. ASC task is designed to predict the sentiment polarity for a particular aspect within a sentence. Because ASC task is the focus of this paper. And in the method, we use the PTMs. Accordingly, we separate our discussion of related work into two areas: Firstly, methods and related research on ASC tasks in recent years. Secondly, development and application of PTM in Aspect-Based Sentiment Analysis tasks.

### 2.1　　Aspect Sentimental Classification

Aspect Sentimental Classification is another important subtask of ABSA. Generally, the aspects can be instantiated as aspect terms or aspect categories, resulting in two ASC problems: aspect term-based sentiment classification and aspect category-based sentiment classification. In fact, the main research issues of these two subtasks are the same. What they explore is how to use aspects and context to classify sentients at aspect level.

In recent years, ASC based on deep learning has aroused widespread interest. Many ASC models based on neural networks [15][16][17] have been proposed and brought great performance improvements [18]. Neural models such as TC-LSTM [20] were created in order to model the interaction between the aspect and sentence context. Different parts of a sentence have different functions in specific aspects, so attention mechanism is widely used to obtain the representation of specific aspects [21]. As a representative work, Wang et al. [22] purposed Attention-based LSTM with Aspect Embedding (ATAE-LSTM) model. The paper attached aspect embeddedness to each word vector of the input sentence to calculate the attention weight, and can calculate aspect specific sentence embeddedness accordingly to classify sentient. And then came the design of more complex attention mechanisms, which purposed to learn better aspect-specific representations. IAN interactively generate the representations for aspect and sentence attention separately, which was purposed by Ma et al. [23]. Besides, other network structures like the gated network [24] also had a good effect. Recently, the development of preprocessing models has greatly improved the performance of tasks. For instance, Sun et al. [25] utilize the ability of BERT by transforming the task as a pair classification problem. There is another method of the ASC research modeling the syntactic structure of the sentence to infer the polarity of the sentient aspects. As the improvement of dependency analysis based on neural network [27][28] in recent years, better parse tree brought significant improvements to the dependent ASC models. Sun et al. [29] and Zhang et al. [30] utilized word dependencies and the syntactical information to model the dependency tree. What they employ is the graph neural network (GNN) [31][32].

Compared with text-level and sentence-level sentient analysis, aspect-level sentient analysis faces new problems. Aspect-level sentiment analysis technology not only analyzes the explicit language expression structure, but also deeply understands the implicit semantic expression. Besides, Aspect-level sentiment analysis needs to determine the context range in which sentiment is expressed for each evaluation aspect. Aspect-level sentiment analysis technology needs to correctly understand the semantic information of the text word-level and sentence-level.

## 2.2 Pre-Training Models

The pre-trained word embeddings, such as Word2Vec [33] and GloVe [34] were used in the conventional neural ABSA models, coupling with the task-specific neural architecture. Compared with early feature-based models, they have a certain effectiveness. But they can't capture complex sentiment dependencies in the sentence because of the context-independent word embeddings. In recent years, pre-trained language models such as BERT [35] and RoBERTa [36] have brought significant improvements on NLP tasks. Given the wealth of knowledge learned during the pre-training phase, simply leveraging such contextualized embeddings has yielded large performance gains. For example, Li et al. [37] tried using several stacked standard prediction layers on PTM for the E2E-ABSA tasks. On the basis of the original Bert model, RoBERTa has proved the following points through experiments: further increasing the number of pre-training data can improve the model effect; extending the pre-training time or increasing the number of pre-training steps can improve the model effect.

For PTMs, most of the current mainstream models use Transformer as a feature extractor. At this stage, the potential of Transformer has not been fully tapped, and there is still a lot of potential to tap. Besides being the backbone of the ABSA model, PTMs can also benefit from handling ABSA tasks in other ways. For example, language modeling tasks used in the pre-training phase of PTM often bring the ability to perform generative data augmentation. Li et al. [38] leveraged PTM to achieve semantic-preserving augmentation in a generative manner, obtaining clear improvements over baseline methods on a range of ABSA tasks.

At present, it is generally believed in the NLP community that PTMs can accurately reflect the semantics of input words [39]. However, the contextual embeddings obtained through the self-attention mechanism, which capture the full word dependencies in the sentence, may be somewhat redundant for the ABSA task. Because ABSA often doesn't need to capture as many dependencies, doing so is wasteful. How to consolidate meaningful sparse relational structures with PLM, or perfect the intrinsic fully connected self-attention, and obtain ABSA-related representations in a more efficient way, deserves more attention and research work.

## 3 DeBERTa-LCF

In this section, we will introduce that our approach the Aspect-Target Sentiment Classification task using a two-step procedure. In the first step, we use the pre-trained model DeBERTa as a basis. We first briefly introduce the model structure of BERT and DeBERTa. In the second step, BERT shared layer is adopted as the embedding layer and feature extractor layer. And integrate it into the local context focusing mechanism. The main framework of DeBERTa-LCF model is shown in Fig. 3.

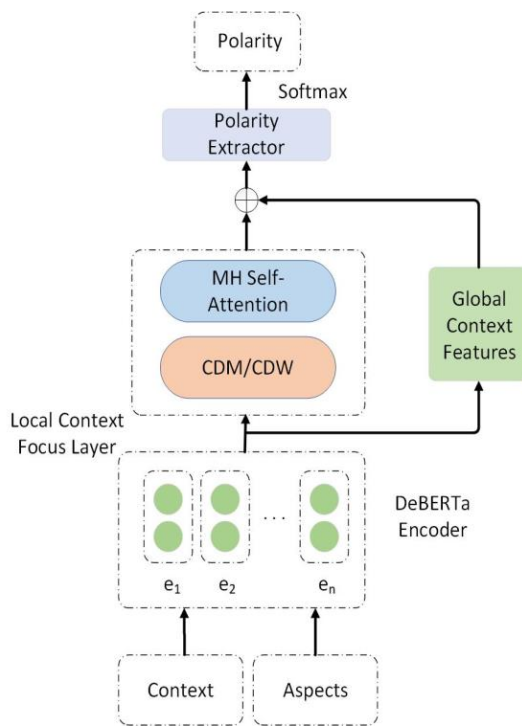

**Fig. 1.** The main framework of DeBERTa-LCF model.

### 3.1 BERT and DeBERTa

BERT and DeBERTa both take Transformers as backbone architecture. The disentangled attention introduced by DeBERTa proposes to separate the content and text position components. Learning attention weights for each component is the main idea of DeBERTa. So it is different from other proposals that sum the position vector to the content vector [40]. This explicit separation allows the model to better segment the positional and content components of the data, where positional embeddings are responsible for generating syntactic features and content embeddings are responsible for semantic features.

$$A_{ij} = \{H_i, P_{i|j}\} \times \{H_j, P_{j|i}\}^T$$

$$= H_i H_j{}^T + H_i P_{j|i}{}^T + P_{i|j} H_j{}^T + P_{i|j} P_{j|i}{}^T \qquad (1)$$

Equation 1 defines the cross-attention matrix used in DeBERTa. In BERT-based models, tokens are represented by content vectors and position vectors. The existing relative position coding methods use a separate embedding matrix to calculate the relative position deviation when calculating the attention weight [41]. It is similar to calculating the attention weight using only the content to content and content to position terms in equation 1.

In ABSA tasks, the goal is to determine the sentiment of each aspect of the entity. Our model utilized DeBERTa pre-trained model parameters to initialize the model for downstream tasks and we apply fine-tuning based on ABSA labeled data to update model parameters. He et al. [42] purposed that without major changes to other parts of the neural network structure, the disentangled attention mechanism can be incorporated into BERT model. We followed the original fine-tuning strategy because these features have been naturally captured by the disentangled attention mechanism.

The disentangled mechanism already took into account the content and relative positions of the context words. So we do not do much fine-tuning on the DeBERTa model. DeBERTa only needs half the data and is better than BERT and RoBERTa. We use the standard softmax layer with categorical cross-entropy loss function, which is the output of the language model, to provides downstream sentiment classification tasks. BERT-Shared layer is regarded as embedded layers, and the fine-tuning process is performed independently according to the joint loss function learned by multi-task.

## 3.2 Local Context Focus

To determine whether the context word belongs to the local context of the targeted aspect, semantic-relative distance (SRD) is purposed for helping models capture local context. SRD is a concept based on token-aspect pairs, which describes the distance between token and aspect. After calculating all the tagged outputs from the attention layer, the output features at each output location above the SRD threshold will be hidden or weakened, while the output features of the local context words will be fully preserved.

We purposed two architectures to focus on local contexts CDM and CDW. Our model focuses on local context by adopting local context focus layer. The input sequence of LCF design is mainly based on the global context. The DeBERTa layer is powerful enough to capture context features. Based on its self-attention mechanism, Multi-Head Self-Attention (MHSA) performs multiple attention functions to calculate the attention score of each contextual word. By MHSA, the features of each code are more closely related to itself. The multiple self-attentions work at the same time, and the obtained results are processed, so that the obtained information is more comprehensive and can be used to extract deep semantic features in the context. MHSA can avoid the negative effects of long-distance context dependence when learning features

## 4    Experiments

### 4.1    Datasets

To evaluate our model, we used the SemEval-2014 and the ACL twitter dataset. There are three datasets for this task: the laptop, restaurants, and twitter reviews. Rest14 and Laptop14 are from SemEval 2014 task 4 [1] containing sentiment reviews from restaurant and laptop domains. Twitter is from Dong et al. [43] which is processed from tweets. In these datasets, users rate their experiences with laptops and restaurants and comments made on Twitter in various aspects. The polarity of each aspect on these datasets may be positive, neutral, and negative. Therefore, the datasets we used contain 12184 total samples and three sentiments (positive, negative, and neutral).
The statistics of these datasets are presented in Table 1.

**Table 1.** Detail of benchmark datasets

|  | Positive | Negative | Neutral |
|---|---|---|---|
| Laptops | 994 | 870 | 464 |
| Restaurants | 2164 | 807 | 637 |
| Twitter | 1561 | 1560 | 3127 |

### 4.2    Compared Models

In order to prove the excellent reliability of the DeBERTa-LCF model described in this paper, it was necessary to show that this model is superior to other models. We compared DeBERTa-LCF with the following state-of-the-art models:
BERT-BASE [44] is the baseline of BERT-based models.
ATAE-LSTM [45] is a classic LSTM based model, which uses attention-based LSTM to explore the relationship between an aspect and sentence content.
GCAE [46] is a CNN based model which based on convolutional neural networks and gating mechanisms.
BERT-ADA [47] is a domain-adapted BERT-based model proposed, which fine-tuned the BERT-BASE model on task-related corpus.
IAN [48] generates the representation of target aspect and context through two LSTM networks respectively, which learns the representation of target aspects and contexts interactively.

RAM [49] is a novel framework based on neural networks to identify the sentiment of opinion targets with a RNN for sentence representation.

**Table 2.** Experimental results of performance.

| Models | Laptop | | Restaurant | | Twitter | |
|---|---|---|---|---|---|---|
| | Acc | F1 | Acc | F1 | Acc | F1 |
| BERT-BASE | 78.52 | 75.5 | 82.12 | 74.53 | 73.4 | 70.22 |
| ATAE-LSTM | 68.7 | 67.23 | 77.1 | 66.84 | 69.22 | 67.4 |
| GCAE | 78.05 | 69.93 | 78.31 | 68.74 | 71.42 | 69.3 |
| BERT-ADA | **79.7** | 75.21 | 81.23 | 72.5 | **74.36** | **71.53** |
| IAN | 72.14 | 70.6 | 78.64 | 70.12 | 70.3 | 67.6 |
| RAM | 74.45 | 71.32 | 80.27 | 71.02 | 69.72 | 67.2 |
| DeBERTa-LCF* | 79.54 | **75.64** | **83.46** | **75.4** | 70.83 | 68.64 |

The results with "*" are derived from our model. We highlight the best results on bold.

### 4.3 Analysis of Overall Performance

Table 2 shows an overview of the experimental results using ACC and F1 metrics for the Laptop, Restaurant and Twitter datasets. The experiments result on the most commonly used the laptop and restaurant datasets of SemEval-2014 and the ACL twitter dataset show that LCF mechanism with DeBERTa has been improved in different degrees, especially in restaurant datasets.

In Restaurant dataset, our proposed DeBERTa-LCF model achieves the best results in terms of both macro-F1 scores and accuracy scores. It gets an improvement of 1.34, 0.87.

In Laptop dataset, our model has also been improved to some extent on macro-F1 scores. And its accuracy scores are only slightly worse than the base-lines.

In Twitter dataset, our model is second only to the best one at present, and the difference is very slight. The comparisons with BERT-BASE models suggest that the local context focus mechanism competent to discover unknown aspects and predict sentiment polarity.

## 5 Conclusion

In this paper, we propose a DeBERTa model with LCF mechanism for aspect-based sentiment classification tasks. We introduce LCF mechanism, which is of great significance for aspect item extraction. LCF designs focus on the local context and learn global context representations in parallel. At the same time, we introduce the DeBERTa model, which is the latest pre-training model. This greatly increased the performance of the model. We applied the DeBERTa model and integrate it and LCF mechanisms for the first time. We conduct a set of experiments on three datasets. The results prove that our model achieves certain performance.